\documentclass[sigconf, screen]{acmart}
\AtBeginDocument{%
  }

\setcopyright{acmlicensed}
\copyrightyear{2025}
\acmYear{2025}
\acmDOI{XXXXXXX.XXXXXXX} % NOTE
\acmISBN{978-1-4503-XXXX-X/2018/06} % NOTE
\acmConference[MM ’25]{33rd ACM International Conference on Multimedia}{October 27–31, 2025}{Dublin, Ireland}
\acmBooktitle{Proceedings of the 33rd ACM International Conference on Multimedia (MM ’25), October 27–31, 2025, Dublin, Ireland}
\acmPrice{15.00}


\usepackage{algorithm}
\usepackage{algpseudocode}
\usepackage{multirow}
\usepackage{booktabs}
\usepackage{bbding}
\usepackage{pifont}
\usepackage{utfsym}
\usepackage{amsmath}
\usepackage{makecell}
\usepackage{array}
\usepackage{amsmath}
\usepackage[inline]{enumitem}
\usepackage{graphicx}
\usepackage{adjustbox}
\usepackage{csquotes}
\usepackage{newfloat}
\usepackage{listings}
\usepackage{bm}
\usepackage{graphicx}
\usepackage{subcaption}
\usepackage{makecell}
\usepackage{threeparttable}
\usepackage{hyperref}

\usepackage{amsthm}

\usepackage{diagbox}

\usepackage{caption}
\usepackage{subcaption}

\begin{document}

\title{Fine-Grained Zero-Shot Object Detection}

\author{Hongxu Ma}
\authornote{Both authors contributed equally to this research.}
\affiliation{
  \institution{Fudan University}
  \city{Shanghai}
  \country{China}}
\email{hxma24@m.fudan.edu.cn}

\author{Chenbo Zhang}
\authornotemark[1]
\affiliation{
  \institution{Fudan University}
  \city{Shanghai}
  \country{China}}
\email{cbzhang21@m.fudan.edu.cn}

\author{Lu Zhang}
\affiliation{
  \institution{Fudan University}
  \city{Shanghai}
  \country{China}}
\email{l_zhang19@fudan.edu.cn}

\author{Jiaogen Zhou}
\authornote{Corresponding author.}
\affiliation{
  \institution{School of Geography and Planning, Huaiyin Normal University}
  \city{Huaian}
  \country{China}}
\email{zhoujg@hytc.edu.cn}

\author{Jihong Guan}
\affiliation{
  \institution{Tongji University}
  \city{Shanghai}
  \country{China}}
\email{jhguan@tongji.edu.cn}

\author{Shuigeng Zhou}
\authornotemark[2]
\affiliation{
  \institution{Fudan University}
  \city{Shanghai}
  \country{China}}
\email{sgzhou@fudan.edu.cn}

\begin{abstract}
  Zero-shot object detection (ZSD) aims to leverage semantic descriptions to localize and recognize objects of both seen and unseen classes. Existing ZSD works are mainly coarse-grained object detection, where the classes are visually quite different, thus are relatively easy to distinguish. However, in real life we often have to face fine-grained object detection scenarios, where the classes are too similar to be easily distinguished. For example, detecting different kinds of birds, fishes, and flowers. 
  In this paper, we propose and solve a new problem called \textbf{F}ine-\textbf{G}rained \textbf{Z}ero-\textbf{S}hot Object \textbf{D}etection (\textbf{FG-ZSD} for short), which aims to detect objects of different classes with minute differences in details under the ZSD paradigm. We develop an effective method called MSHC for the FG-ZSD task, which is based on an improved two-stage detector and employs a multi-level semantics-aware embedding alignment loss, ensuring tight coupling between the visual and semantic spaces. Considering that existing ZSD datasets are not suitable for the new FG-ZSD task, we build the first FG-ZSD benchmark dataset FGZSD-Birds, which contains 148,820 images falling into 36 orders, 140 families, 579 genera and 1432 species. Extensive experiments on FGZSD-Birds show that our method outperforms existing ZSD models. 
\end{abstract}


\maketitle

\section{Introduction}

\begin{figure*}[t]
  \centering
   \includegraphics[scale=0.46]{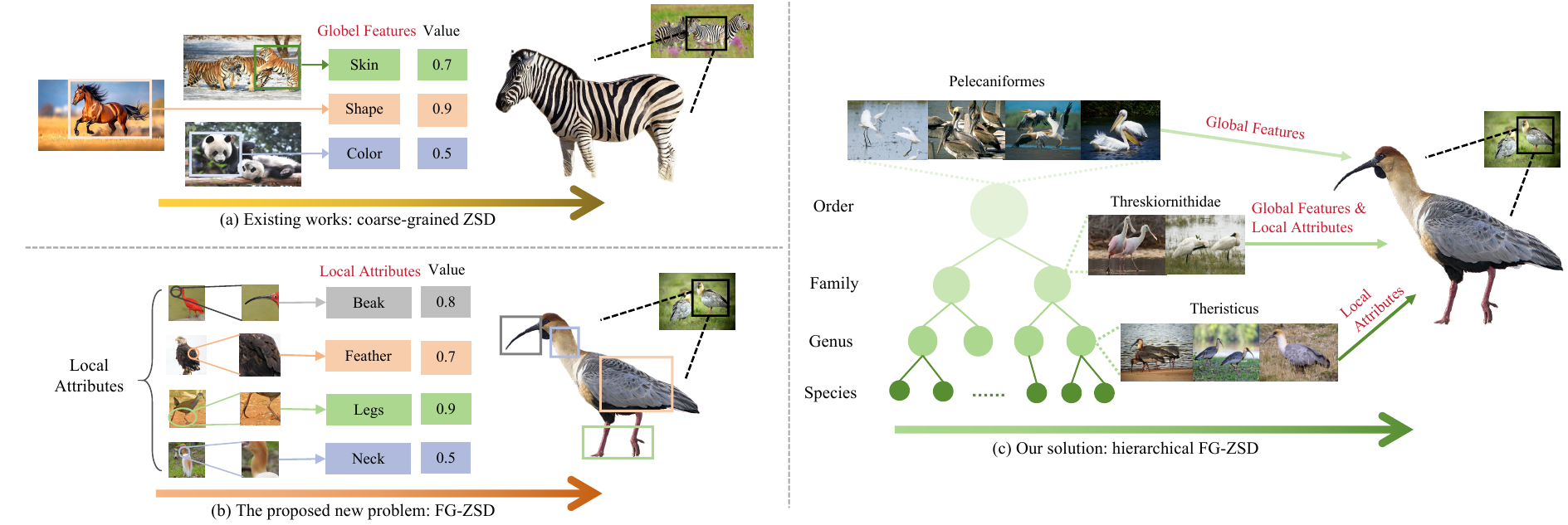}
   \caption{Comparison between this paper and existing works. (a) Existing works: coarse-grained ZSD, where the categories are generally coarsely split and easy to distinguish. Objects are detected based on the global relationships of features between categories.
   (b) The proposed new problem: FG-ZSD, where the categories are split at a fine granularity and difficult to distinguish. To detect objects, the subtle differences of local attributes between categories should be properly exploited. (c) Our solution: hierarchical FG-ZSD, which exploits both higher-level global features and lower-level attributes to detect objects from top categories to bottom categories hierarchically.}
   \label{fig:motivation}
\end{figure*}

In the past decade, generic object detection~\cite{ren2015faster} has achieved great progress in the computer vision area. Usually, large amounts of labeled training data are required to train the detector~\cite{zhang2024musetalk, zhang2024pixelfade,ma2025generativeregressionbasedwatch}, which can recognize only objects of the categories that appear in the training data. However, in many real scenarios, samples of certain categories are difficult or even impossible to obtain, such as endangered species in the wild. Therefore, in recent years few-shot object decetion~(FSOD)~\cite{fan2020few,han2022few} and zero-shot object detection~(ZSD) \cite{bansal2018zero,rahman2018zero,demirel2018zero, jiang2025multimodal} have received increasing attention from academia and industry. This paper focuses on zero-shot object detection, which aims to locate and recognize objects of unseen classes via capturing and transferring knowledge from seen classes to unseen classes. 

Existing ZSD works are mainly \textit{coarse-grained} object detection, where the categories are visually quite different, i.e., coarsely split, thus are relatively easy to  
distinguish. These 
methods learn to 
detect unseen objects based on the global relationships of features between seen and unseen categories, as shown in Fig.~\ref{fig:motivation}(a). Here, a zebra is detected based on its shape similarity (or value) to the horse, skin similarity to the tiger, and color similarity to the panda. And the shape/skin/color similarity is evaluated globally between objects of different categories.
However, in the real world there are many scenarios where we have to detect objects of different categories that are visually quite similar, i.e., the categories are split at a fine granularity, so difficult to distinguish. For example, it is difficult to distinguish American flamingo from Andean flamingo even though detailed descriptions are provided. In such fine-grained scenarios, to detect objects, the subtle differences of local object attributes between different categories should be properly exploited as in fine-grained image classification~\cite{he2019and}.

In this paper, we propose a new problem of \textbf{F}ine-\textbf{G}rained \textbf{Z}ero-\textbf{S}hot object \textbf{D}etection, \textbf{FG-ZSD} for short, which aims to perform zero-shot object detection at a fine-grained class level, as shown in Fig.~\ref{fig:motivation}(b) .  Obviously, a FG-ZSD model should be able to capture not only the shared features but also the subtle differences in local object attributes of different fine-grained classes. Therefore, FG-ZSD is more challenging yet practical than ZSD. 
To solve the FG-ZSD problem, we face two challenges:

The first challenge is \textbf{dataset}. Though there are a number of datasets for fine-grained tasks~\cite{zhuang2018wildfish,yu2021ap,maji2013fine,coates2011analysis,wah2011caltech,nilsback2008,krause20133d,bossard2014food}, they are mainly for classification tasks, not suitable for detection tasks\footnote{CUB~\cite{wah2011caltech} includes bounding boxes but contains only single object per image, limiting it to classification tasks}. Besides, they are simply annotated with flat class labels and do not support complex tasks due to the lack of class and attribute descriptions, and hierarchical information, etc. Even for the current mainstream ZSD datasets such as VOC~\cite{voc} and COCO~\cite{coco}, they have a simple one-level flat class structure with only 80 and 20 classes respectively, and the categories are relatively coarse, which are not suitable (or too easy) for the FG-ZSD task. 

The second challenge is \textbf{methodology}. When applied to fine-grained scenarios, existing ZSD methods have two main drawbacks: 1) most ZSD methods learn visual and semantic feature mappings to achieve alignment~\cite{bansal2018zero,demirel2018zero,gupta2020multi,mao2020zero,li2025vista,yan2022semantics,zheng2020background,li2025visuals}. Since the unseen classes are invisible during training and the classes are too similar to each other, it is easy to bias the seen classes in inference, recognizing the unseen classes as background or seen classes. 2) Though some ZSD methods synthesize visual features from semantic vectors ~\cite{gen1,gen2,gen3,gen4,gen5,sarma2022resolving}, they consider only information of category labels and the whole sentences of category description, while ignoring discriminative fine-grained information at word level. 

To address the challenges above, 
on the one hand, we construct the first large, high-quality wild-bird dataset named \textbf{FGZSD-Birds}.
FGZSD-Birds has totally 148,820 images with high-quality labeled bounding boxes, containing 36 orders, 140 families, 579 genera and 1432 species, and covering more than 90\% of the world's waterbirds. For each finest (i.e., the lowest level) category, we provide not only the category hierarchical information of order, family, genus and species from Wikipedia, but also the textual descriptions of the category features, %which are carefully edited and checked. 
And each image has not only manually annotated and cross-checked category labels and bounding boxes, but also 3-sentence captions.
%multi-language
In addition to FG-ZSD, FGZSD-Birds can actually also support dozen of other downstream tasks, including  few-shot/zero-shot/fine-grained/hierarchical  classification and object detection, image captioning/generation etc. Moreover, considering the difficulty of obtaining samples of endangered birds and %the high demand for expert annotations
high cost of annotations, FGZSD-Birds is of great importance in endangered bird conservation.
%Note that 

On the other hand, we propose a novel FG-ZSD method named \textbf{M}ulti-level \textbf{S}emantics-aware and \textbf{H}ierarchical \textbf{C}ontrastive 
model (\textbf{MSHC} for short). 
First, we develop an attention-based visual-semantic similarity loss that trains text/image encoder to achieve alignment of visual and semantic features at the sentence and word level. 
Second, we devise a multi-level semantics-aware generative module for unseen classes, which can synthesize fine-grained details of different sub-regions of an image by focusing on the related words in the text description. 
Specifically, it applies channel scaling and shift operations to visual feature maps, and uses multi-level semantic vectors for high-diversity feature synthesis via multi-stage refinement.
Last, as shown in Fig.~\ref{fig:motivation}(c), to 
combine higher-level global features and lower-level attributes to enhance the model's ability to distinguish fine-grained categories, we design a multi-level semantics-aware hierarchical contrastive learning loss, which constrains the feature space to be consistent with the class hierarchy.
Note that although FGZSD-Birds dataset focus on birds for our FG-ZSD study, our method MSHC is a general solution for FG-ZSD NOT limited to any specific fine-grained categories and can seamlessly adapt across diverse fine-grained datasets.

\begin{table*}[t]
  \centering
  \caption{Dataset Comparison. FGZSD-Birds is the only large-scale and high-quality fine-grained benchmark, provides category labels, bounding boxes,  image captions and hierarchical information, supports FG-ZSD and other tasks.}
  \resizebox{0.9\textwidth}{!}{
  \begin{tabular}{@{}cccccccc@{}}
    \toprule
    \multirow{2}{*}{Dataset} & \multirow{2}{*}{\#Images} & \multirow{2}{*}{\shortstack{\#Classes at \\the lowest level}}& \multirow{2}{*}{\shortstack{Hierarchical structure}}& \multicolumn{4}{c}{Some supported tasks} \\
    \cmidrule(l){5-8}
    & & & & \textbf{FG-ZSD}& Image classificaiotn& Image captioning& Text to image generation \\
    \midrule
    FGVC-Aircraft~\cite{maji2013fine}   & 10.2K& 102& \checkmark& -& \checkmark & - & - \\
    STL-10~\cite{coates2011analysis}    & 13K&             10& & -& \checkmark& -  & - \\
    CUB~\cite{wah2011caltech}           & 11.8K& 200& \checkmark& -& \checkmark& \checkmark & \checkmark \\
    WildFish~\cite{zhuang2018wildfish} & 3K& 44& -& -& \checkmark& - & - \\
    Oxford Flower~\cite{nilsback2008}   & 8.2K& 102& -& -& \checkmark& - & - \\
    AP-10K~\cite{yu2021ap} &  10K& 54& \checkmark& -& \checkmark& - & -  \\
    Stanford Cars~\cite{krause20133d}   & 16.2K& 196& -& -& \checkmark& - & - \\
    Food-101~\cite{bossard2014food}      & 101K& 101& -& -& \checkmark& - & - \\
    %FGVD~\cite{khoba2022fine}   & 210           & \checkmark   & 5.5K & \checkmark & \checkmark & - & - \\
    \textbf{FGZSD-Birds~(Ours)}  & \textbf{148.9K}& \textbf{1432}& \Checkmark& \Checkmark & \Checkmark & \Checkmark & \Checkmark \\
    \bottomrule
  \end{tabular}
  }
  \label{tab:dataset comparisons}
\end{table*}

Our contributions are as follows:
    1) We propose a new ZSD problem called fine-grained zero-shot object detection (FG-ZSD in short), which is more challenging yet practical than ZSD. 
    2) We develop the first FG-ZSD method MSHC, which uses multi-level aligned semantic vectors to generate unseen class images and multi-level semantics-aware hierarchical contrastive learning to constrain the feature space. 
    3) We construct the first large-scale, high-quality FG-ZSD dataset FGZSG-birds, which can support not only FG-ZSD, but also dozen of other tasks. 
    4) We conduct extensive qualitative and quantitative experiments on FGZSD-Birds, which show that the proposed method outperforms existing SOTA ZSD models.

\section{Related Work}

\subsection{Zero-shot Object Detection~(ZSD)}
Existing ZSD methods can be roughly divided into two groups: alignment-based methods and generation-based methods. The former \cite{bansal2018zero,demirel2018zero,gupta2020multi,mao2020zero,rahman2020zero,yan2022semantics,yan2020semantics,zheng2020background} learns an embedding function that maps visual features to semantic features or maps visual and semantic features to a common space to achieve alignment, and then uses nearest neighbor search to obtain classification results. The latter \cite{gen1,gen2,gen3,gen4,gen5,sarma2022resolving, li2023zero,huang2024m,wang2024saui} converts the ZSD problem into a 
generic object detection, which trains a generative model with semantic and visual features of the seen classes, then synthesizes visual features of the unseen classes that will be trained together with the seen classes to get a model to recognize both the seen and unseen classes. However, existing ZSD methods are mainly coarse-grained object detection, which cannot be directly applied to the FG-ZSD problem proposed in this paper. In this paper, we build a large-scale, high-quality benchmark dataset and try to combine the advantages of both types to develop an effective method.

% Existing ZSD methods can be categorized into alignment-based and generation-based approaches. Alignment-based methods \cite{bansal2018zero,demirel2018zero,gupta2020multi,mao2020zero,rahman2020zero,yan2022semantics,yan2020semantics,zheng2020background} learn an embedding function to map visual features to semantic features or to a common space, followed by nearest neighbor search for classification.  The latter \cite{gen1,gen2,gen3,gen5,li2023zero,huang2024m,wang2024saui} converts the ZSD problem into a generic object detection, which trains a generative model with semantic and visual features of the seen classes, then synthesizes visual features of the unseen classes that will be trained together with the seen classes to get a model to recognize both the seen and unseen classes. However, these methods are typically for coarse-grained object detection and are not directly applicable to the FG-ZSD task as proposed here. 
% % We address this by building a high-quality benchmark and combining the strengths of both approaches.

\subsection{Fine-grained Image Recognition (FGIR)}
FGIR aims to differentiate the sub-categories under a broader category, such as distinguishing the species of birds. Existing FGIR methods roughly fall two groups: 1) part-based methods~\cite{he2019and,he2018fast,wang2020weakly,dubey2018maximum,xu2024enhancing} that first detect discriminative regions in images, then perform fine-grained classification by feature extraction. 2) Attention-based methods~\cite{zheng2017learning,sun2020fine,chang2021your,rao2021counterfactual，zhang2024learning} that harness the power of attention mechanisms to automatically learn discriminating features. 
There are also some fine-grained zero-shot learning works. S2GA~\cite{ji2018stacked} is a novel stacked semantics-guided attention model, Huynh et al.~\cite{huynh2020fine} designed a dense attribute-based attention mechanism to align attribute embedding. However, all these methods above perform classification under a flat category structure. %, which overlooks the hierarchical category structure existing widely in real life. 
%Since the differences between categories are too small, it is not easy to distinguish different categories especially when the number of categories is too large.%
Differently, Chang et al.~ \cite{chang2021your} proposed a hierarchical taxonomy for the FGIR problem, they 
%focuses on a redefinition of FGIR and 
simply stacked the features of different layers to optimize classification. In this paper, we propose a multi-level semantics-aware hierarchical contrastive learning to make the feature space consistent with the hierarchical taxonomy so that the FG-ZSD model can learn better features.

\subsection{Hierarchical image recognition (HIR)}
HIR is to classify images under a hierarchical class structure. HPL~\cite{zhang2019hierarchical} is a hierarchical prototype learning method for zero-shot learning. 
HSVA~\cite{chen2021hsva} devised a hierarchical visual semantic alignment method but does not account for the inherent hierarchical structure in the data.
HiCLPL~\cite{zhang2022hierarchical} is a hierarchical few-shot object detection method based on hierarchical contrastive learning and probabilistic loss. These works neglect the semantic distance in the class hierarchy, thus cannot make the feature space consistent with the hierarchical taxonomy well.

\subsection{Text-to-image Generation (T2I)}
T2I tries to translate textual descriptions into corresponding visual content. Existing T2I methods can be broadly divided into two types: GAN-based and large model-based. Recent large-scale pretrained autoregressive and diffusion models~\cite{ramesh2021zero, rombach2022high, mi2025data, xu2025hunyuanportrait, lu2024coarse} have shown excellent capability in generating intricate scenes. However, they do not provide a smooth latent space to align visual features and semantic vectors, and multi-step generation disrupts the synthesis process and the latent space, which is not suitable for fine-grained scenes. In GAN-based methods, GAN-INT-CLS~\cite{reed2016generative} is the first to use conditional GAN to generate images. Subsequent models such as StackGAN~\cite{zhang2017stackgan,zhang2018stackgan++}, AttnGAN~\cite{xu2018attngan}, and DM-GAN~\cite{zhu2019dm} design a hierarchical structure of GANs and loss functions to enhance the image resolution, but this structure makes the generated images look like simple combination of visual features at different scales. DF-GAN~\cite{tao2022df} is a simplified method to synthesize high-resolution images in one step but does not utilize word-level information, which limits the ability to synthesize fine-grained visual features.

\section{FG-ZSG Benchmark}
\label{sec:benchmark}
We construct the first benchmark dataset for the FG-ZSD task, which is called \textbf{FGZSD-Birds}. FGZSD-Birds sets a  standard for FG-ZSD model evaluation, and lays down the foundation for further exploration of FG-ZSD techniques.

\subsection{Data Construction} The source of our data consists of two parts: most of it comes from copyright-free websites, and the rest from camera shots. 
To ensure the quality of the dataset, we carefully filter all data to remove poor quality (especially blurred) and duplicate images. Before annotating, we provide rigorous training to the annotators on the external/attribute features, body structure, and hierarchical information of each species, and select skilled annotators based on the results of the training. The annotators need to not only annotate all objects with bounding boxes and labels, but also provide captions for each image and class descriptions for each class from Wikipedia. We assign three annotators to each image and ask them to annotate the images using a consensus-based approach, with a final cross-check to ensure accuracy and high quality of annotations. 

\subsection{Seen/unseen Split} 
\begin{figure}[t]
  \centering
   \includegraphics[width=0.7\linewidth,height=0.16\textheight]{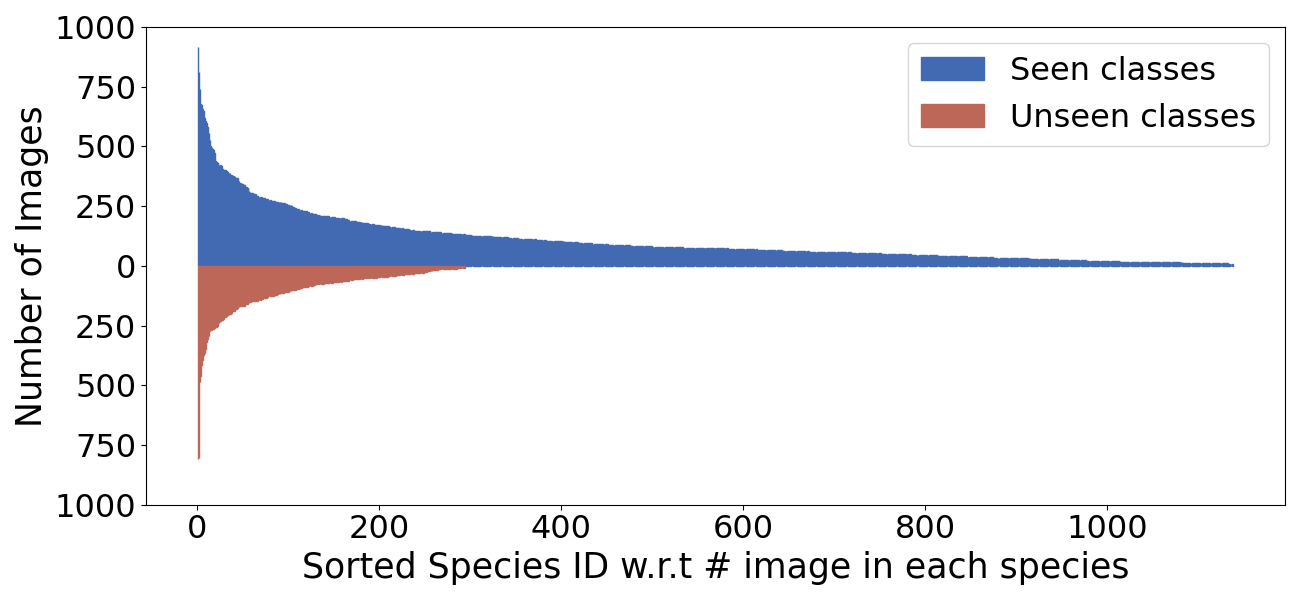}
   \vspace{-1em}
   \caption{The distribution of images for seen and unseen species, consistent with a long-tailed distribution.}
   \label{fig:specimgs}
   \vspace{-1.4em}
\end{figure}

To perform FG-ZSD, we need to divide the classes into seen and unseen classes. In our split strategy, we ensure that each genus with more than 1 species has at least an unseen species. Eventually, there are 1,138 seen classes and 294 unseen classes, with a ratio of about 4:1. Overall, our seen/unseen class split has the following advantages: 1) The use of genus-centered division allows for a more comprehensive coverage of the avian taxonomy and enables the model to leverage more shared attributes for effective FG-ZSD. 2) The fact that unseen classes come from different genera increases interclass variation within unseen classes, which in turn improves the model's discriminative and generalization power. 3) As shown in Fig.~\ref{fig:specimgs}, the long-tailed distribution of the number of images for both seen and unseen classes indicates that our dataset is consistent with the real distribution of birds in the wild. %, demonstrating its practicality in real scenarios. % and potential value in conservation of endangered birds. 

\subsection{Comparison and Statistics} 
To demonstrate the uniqueness and importance of our dataset, in Tab.~\ref{tab:dataset comparisons} we compare FGZSD-Birds with eight existing fine-grained datasets. We can see that FGZSD-Birds has the largest size and the most number of classes at the finest/lowest level. In addition to FG-ZSD, it also supports other tasks, including classification, detection, captioning, and generation.   Tab.~\ref{tab:Data} presents the major statistical results of FGZSD-Birds, which contains 148,820 images with labels and high-quality bounding boxes, 446,460 image captions, and 1,432 category descriptions of 1,432 bird species, which are further grouped into 579 genera, 140 families and 36 orders, covering more than 90\% of the world's waterbirds. 

\section{Methodology}
\subsection{Problem Definition}
FG-ZSD is an extension of ZSD, which aims to locate and recognize objects of unseen classes at the fine-grained class level. Formally, we have two disjoint sets of classes (the classes at the lowest level in the class hierarchy): seen class set $C_{s}$ and unseen class set $C_{u}$ where $C = C_{s}\cup C_{u}$ and $C_{s} \cap C_{u}=\varnothing$. The training set contains only images of seen classes $C_{s}$, where each image contains the bounding box coordinates and the corresponding class labels. Corresponding to ZSD and GZSD (Generalized ZSD), we also consider two settings: FG-ZSG and FG-GZSD. In the FG-ZSD setting, the testing set contains only images of unseen classes $C_{u}$, while in the FG-GZSD setting images of both the seen and unseen classes are included in the testing set. Meanwhile, during the training and testing phases, category descriptions and image-text pairs are provided for both seen and unseen classes. 

\subsection{Model Framework} 
\begin{figure*}[t]
  \centering
   \includegraphics[width=0.8\linewidth]{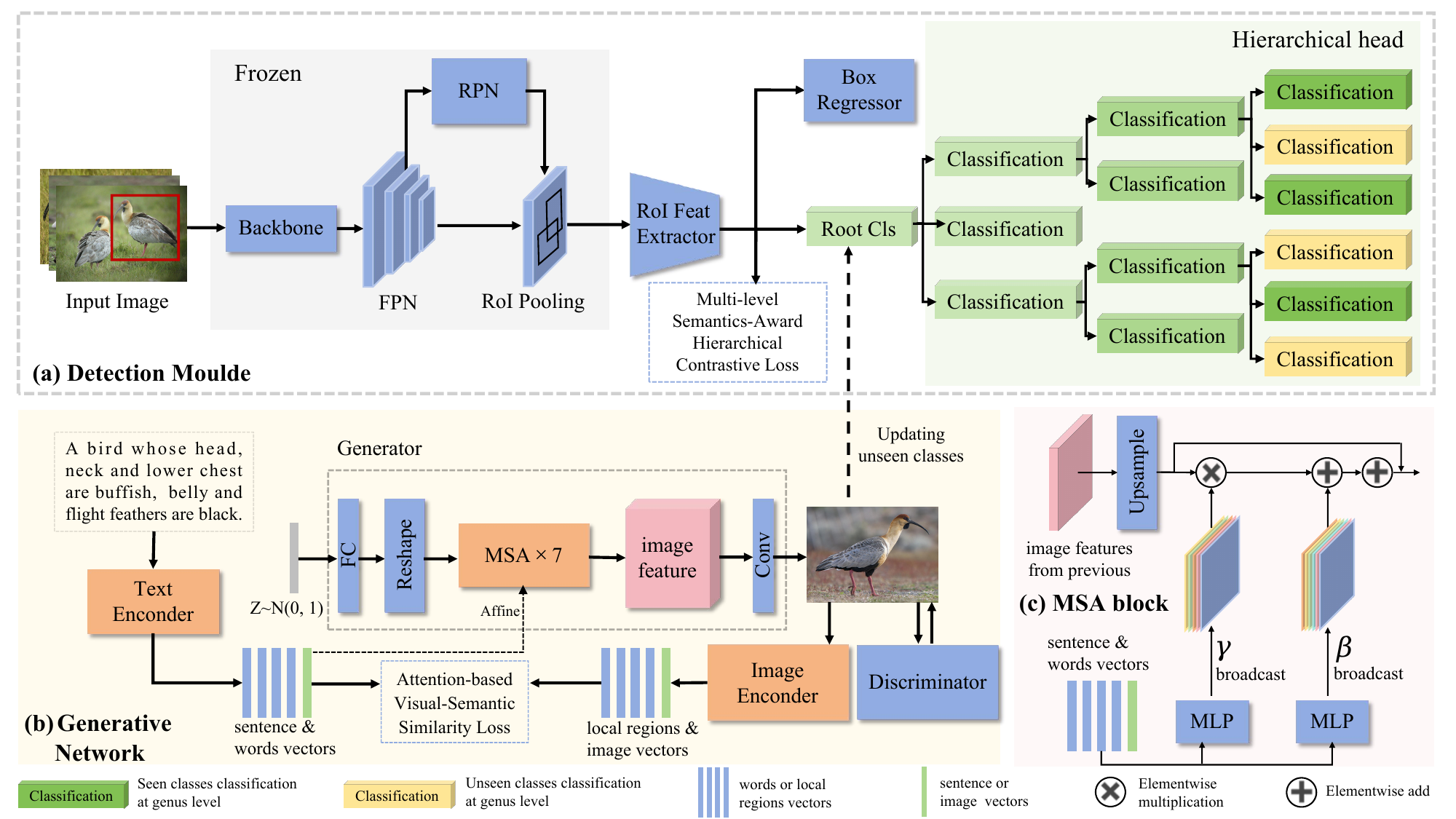}
   \caption{The framework of our method MSHC, which consists of mainly (a) a detection module employs a hierarchical classification head with the same hierarchical structure as the FGZSD-birds dataset, and a multi-level semantics-aware hierarchical contrastive learning loss to make the feature space consistent with the hierarchical taxonomy. (b) a Multi-level Semantics-Aware (MSA) generative network that synthesizes fine-grained details of different sub-regions of an image by focusing on the related words in the text description of unseen classes, where a core part is (c) the Multi-level Semantics-Aware (MSA) block.}
   \label{fig:framework}
   % \vspace{-1em}
\end{figure*}

\begin{figure}[t]
  \centering
   \includegraphics[scale=0.4]{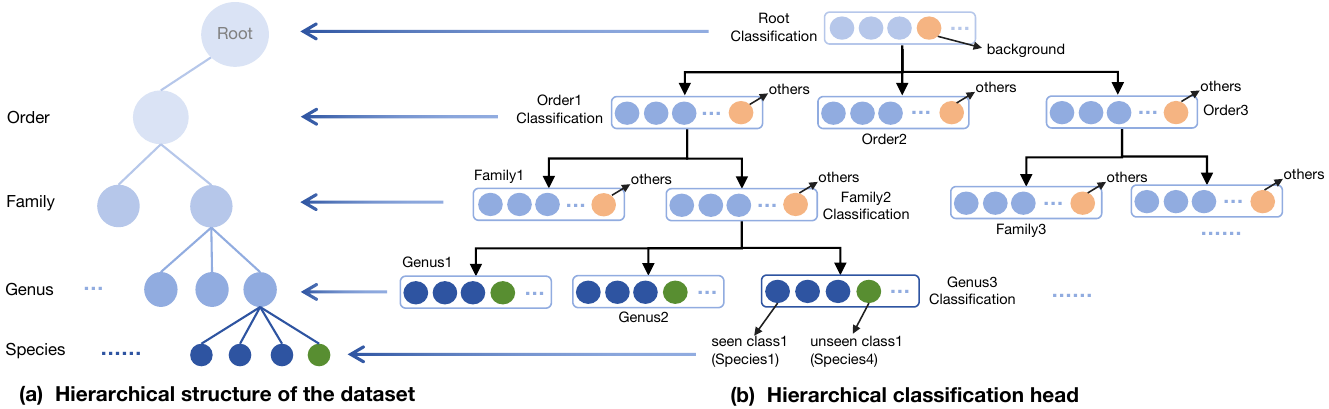}
   \caption{Illustration of the correspondence between the hierarchical classification head and the dataset class hierarchy.}
   \label{fig:hierarchical_cls_head}
\end{figure}

Fig.~\ref{fig:framework} illustrates the framework of our method MSHC, which consists of two major components: an object detection module shown in Fig.~\ref{fig:framework}(a), and a generation module called Multi-Level Semantics-Aware Generative Network (MSA-Generative Network) shown in Fig.~\ref{fig:framework}(b). The object detection module is based on Faster-RCNN~\cite{girshick2015fast}, 
but replaces the original classification head with a hierarchical classification head, which has the same hierarchical class structure as the FGZSD-birds dataset as shown in Fig.~\ref{fig:hierarchical_cls_head}.

\begin{table}[t] 
  \centering
  \caption{Statistics of the FGZSD-Birds dataset.}
  \resizebox{0.35\textwidth}{!}{
  \begin{tabular}{@{}llll@{}}
    \toprule
     & Total & Seen classes & Unseen classes \\
    \midrule
    % \#classes(\#cls) & 1432 & 1138 & 294 \\
    \#orders & 36 & 36 & 28 \\
    \#families & 140 & 140 & 100\\
    \#genera & 579 & 578 & 294 \\
    \#species(\#cls) & 1432 & 1138 & 294\\
    \#images(\#img) & 148820 & 119332 & 29488 \\
    Range of \#img/\#cls &[13, 1325] & [30, 1325] & [13, 897] \\
    Avg of \#img/\#cls & 103.92 & 104.86 & 100.30 \\
    % \#boxes & 151265 & 121273 & 29992 \\
    \#img captions & 446460 & 357996 & 88464\\
    box-size range & [5, 9559765] & [5, 9559765] & [5, 9235740]\\
    box W/H range & [0.25, 9.15] & [0.25, 9.15] & [0.32, 6.37]\\
    \bottomrule
  \end{tabular}
  }
  \label{tab:Data}
\end{table}

Specifically, the level of hierarchy is assigned from 0 (root) to $L$ (the lowest level, corresponding to leaf nodes). In the hierarchical tree $H_t$, each class $c_i\in C$ (classes at the lowest level) corresponds to a leaf node $N_{c_{i}}^L$, and there is only one path $P_{c_i}$ from the root node $N_{c_{i}}^0$ to each leaf node. Each non-leaf node $N_{c_{i}}^{l}$ in the path $P_{c_i}$ corresponds to a classifier $F_{c_{i}}^{l}$, where $l \in [0, L-1]$ and each $F_{c_{i}}^{l}$ needs only to distinguish the more fine-grained subclasses of $N_{c_{i}}^{l}$. The regression head uses class-agnostic regression and is set only at the root node to reduce parameters.

First, we train the hierarchical Faster-RCNN with seen class data, after which we freeze the backbone and RPN parameters. Then, we train the generative network with the aligned visual-semantic vectors to learn the mapping relationships between texts and images, and use the generator to generate fine-grained images for the unseen classes as shown in Fig.~\ref{fig:framework}(b), where a core part is the Multi-level Semantics-Aware (MSA) block illustrated in Fig.~\ref{fig:framework}(c).  Finally, these synthesized images are used to fine-tune the hierarchical Faster-RCNN previously trained by the seen classes to develop a new detector for the FG-ZSD task. 
The core of our method is to generate images of unseen classes at a fine-grained level and effectively recognize the unseen objects. Specifically, we uses a multi-level aligned semantic vectors to synthesize fine-grained details of different sub-regions of an unseen image, and a multi-level semantics-aware hierarchical contrastive loss to make the feature space be consistent with the hierarchical taxonomy.

\subsection{MSA-Generative Network}
Fig.~\ref{fig:framework}(b) shows the structure of the MSA-Generative Network, which consists of a visual-semantic alignment learning module that adopts a attention-based visual-semantic similarity loss, a single-stage generator with seven MSA blocks for incrementally enhancing the image details and refining the visual features, and a discriminator. 
Its major parts are elaborated in detail below.

\subsubsection{Visual-semantic alignment.} 
The visual-semantic alignment module facilitates the training of a text/image encoder through the alignment of sub-regions within images with corresponding words in textual content, with the concomitant assessment of image-text similarity at the granularity of individual words.
% Specifically, w
We respectively encode textual descriptions and images into $e\in \mathbb{R}^{D \times N_{text}}$ and $v\in \mathbb{R}^{D \times N_{img}}$ using a bi-directional LSTM as a text encoder and a CNN as an image encoder. Here, $D$ is the vector dimension, $N_{text}$ and $N_{img}$ (fixed to 289) represent the number of words in the text and image sub-regions,~respectively. Then the similarity matrix $\bar{S} = e^T v$ of image sub-regions and words can be normalized by:
\begin{equation} 
   S_{i,j} = \frac{exp(\bar{S}_{i,j})}{\sum_{k=1}^{N_{img}}exp(\bar{S}_{i,k})} 
  \label{eq:1}
\end{equation}
where the $S_{i,j}$ represent the similarity between the $i_{th}$ word and the $j_{th}$ sub-region in the image. So the similarity between the $i_{th}$ word $e_i$ and the image is computed based on the attention mechanism with the following equations:
\begin{equation} 
\label{eq:2}
    R(r_{i},e_{i}) = \frac{r_{i}^{T}e_{i}} {||r_{i}||\ ||e_{i}||}
\end{equation}
\begin{equation} 
\label{eq:3}
    r_{i} = \sum_{j=1}^{N_{img}} \frac {exp(S_{i,j}) \cdot v_{j}}{\sum_{k=1}^{N_{img}}exp(S_{i,k})}
\end{equation}
where $r_i$ is the region context vector,  %Based on Eq.~(\ref{eq:2}), 
so the similarity between the text ($E$) and the image ($V$) can be difined as:
\begin{equation} 
\small
   Sim{(V, E)} = \log_{}{\left ( \sum_{i=1}^{N_{text}} exp{(\xi R(r_{i},e_{i}))} \right )}^{\frac{1}{\xi}}  
  \label{eq:4}
\end{equation}
where $\xi$ is a hyperparameter that used to determine the influence of the word that matches the image sub-region most closely on the overall match $Sim{(V, E)}$. When $\xi \rightarrow \infty$, it will amplify the importance of the most relevant words to the image, making $Sim{(V, E)} \approx max_{i=1}^{Ntext}R(r_{i},e_{i})$. 

Finally, for a batch of image-text pairs $\{{V_{i}, E_{i}}\}_1^N$, the attention-based visual-semantic similarity loss is 
\begin{equation} 
   \mathcal{L}_{\mathrm{avss}} = \frac{1}{2} \left( \ell(V_i, E_i) + \ell(E_i, V_i) \right )
  \label{eq:5}
\end{equation}
\begin{equation} 
   \ell (V_i, E_i) = - \sum_{i=1}^{N} \log_{} {\left(\frac{exp(Sim{(V_i, E_i))}}{\sum_{k=1}^{N} exp(Sim{(V_k, E_i))}}\right)}
  \label{eq:6}
\end{equation}

We employ the loss function presented in Eq.~(\ref{eq:5}) to train the text/image encoder, 
and the vectors from the encoder will be used in subsequent steps.

\subsubsection{Multi-level semantics-aware block.} The core of the MSA-generative network is the MSA Block, which consists of an upsampling block, two multi-level semantics-aware text-image fusion blocks and a residual link, as shown in Fig.~\ref{fig:framework}(c). Specifically, the input of the MSA consists of the word vectors $e$ and sentence vectors $s$ of the encoded text, and the visual feature maps $f$ generated by the previous MSA block. The upsampling block doubles the resolution of the feature maps through a bilinear interpolation operation. The multi-level semantics-aware text-image fusion block consists of multiple affine transformations and ReLU layers.
% To deepen the text-image fusion process 
The first fusion block consists of two MLPs that predict the channel scaling parameter $\gamma_s$ and shift parameter $\theta_s$ at the sentence level: 
\begin{equation} 
  \gamma_s = MLP_{\gamma_s}\left( s\right),\ \theta_s = MLP_{\theta_s}\left( s\right)
  \label{eq:7}
\end{equation}
where $s$ is semantic vectors of the whole sentence. To further adjust the details of the image sub-regions at each refinement stage, at the word level, we compute the affine transformation parameters for each word as follows:
\begin{equation} 
  \begin{split}
    \gamma_w &= \sum_{i=1}^{N_{text}}R\left(r_f, e_i\right)\cdot MLP_{\gamma_w}\left( e_i\right),\\
    \theta_w &= \sum_{i=1}^{N_{text}}R\left(r_f, e_i\right)\cdot MLP_{\theta_w}\left( e_i\right)
  \label{eq:8}
  \end{split}
\end{equation}
where $e_i$ is the $i_{th}$ word , $r_f$ is the region context vectors computed by Eq.~(\ref{eq:3}) and $R\left(r_f,e_i\right)$  is the similarity by Eq.~(\ref{eq:2}) between the $i_{th}$ word $e_i$ and $f$ as affine transformation weight. 

By prioritizing words that are highly correlated with image features, model can accurately refine the details of the image and deepen the text-image fusion process,
%perform text-image fusion more accurately
resulting in more accurate fine-grained image. The whole formulation of the MSA-Generative Network is as follows:
\begin{equation} 
  \begin{split}
    L_{D} &= -\mathbb{E}_{x\sim P_{r}}[\min(0, -1 + D(x, e))] \\
          &\quad -\frac{1}{2} \mathbb{E}_{G(z)\sim P_{s}}[\min(0, -1 - D(G(z), e))] \\
          &\quad -\frac{1}{2} \mathbb{E}_{x\sim P_{r}}[\min(0, -1 - D(x, \hat{e}))] \\
    L_{G} &= -\mathbb{E}_{G(z)\sim P_{s}}[D(G(z), e)]
  \end{split}
  \label{eq:8-1}
\end{equation}
where $z$ is a normal-distributed noise vector and $z \in \mathbb{R}^{100}$, $e$ is given vectors of both words and sentence while $\hat{e}$ is a mismatched vectors, $P_{s}, P_{r}$ is the distribution of synthetic data, real data respectively.

\subsection{Multi-level Semantics-aware Hierarchical Contrastive Learning}
\begin{figure}[t]
  \centering
   \includegraphics[width=1\linewidth]{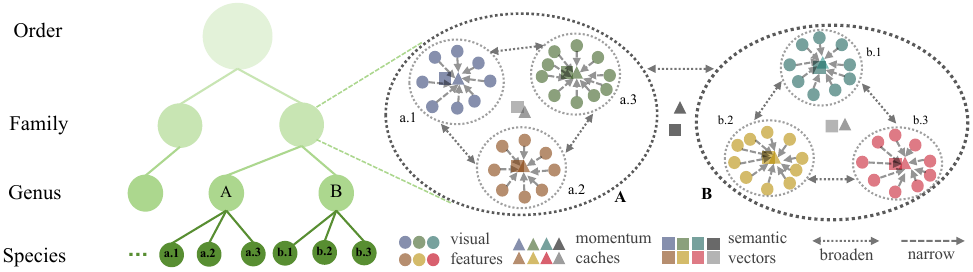}
   \caption{Illustration of multi-level semantics-aware hierarchical contrastive learning. We use inherent semantic distance in the category hierarchy to constrain feature space.}
   \label{fig:SA-hicl}
\end{figure}
Hierarchical semantic structure is naturally present in fine-grained datasets, where subclasses within a parent class share common attributes while possessing unique features.  
As shown in Fig.~\ref{fig:SA-hicl}, the feature space for subclasses within the same parent category should be closely clustered, while distinctly separated from those of different categories. Traditional contrastive learning methods~\cite{zhang2019hierarchical,zhang2022hierarchical} treat all negative classes uniformly, enforcing an equal distance between the positive class features and those of any negative class. This approach neglects the varying semantic distances that naturally exist within the hierarchical structure of categories. For example, the distance between American flamingo and Andean flamingo in the feature space should be less than the distance between them and the woodpecker. 
To effectively utilize the rich semantic information, we design a multi-level semantics-aware hierarchical contrast loss to constrain the feature space and a momentum cache mechanism to enhance the representation consistency.
Meanwhile, we propose a momentum update strategy to continuously refine these representations.

\subsubsection{Multi-level semantics-aware hierarchical contrastive loss.} Each node $N_{c_i}^l$ in the hierarchical tree $H_t$ has a prototype, which is stored by a momentum cache % (momentum buffer,  cache, buffer, )
$\theta_{c_i}^l$. %where $l \in [0, L]$ and $c_i$ is the ground-truth label of a feature map $x$. 
Our goal is to maximize the consistency between feature $x \in c_i$ and {$\{\theta_{c_i}^l|l \in [0, L]\}$}, and minimize the consistency between  $x \in c_i$ and {$\{\theta_{C^{\setminus c_i}}^l|l \in [0, L]\}$} based on semantic distances between $c_i$ and {$\{C^{\setminus c_i}\}$}. Specially, for a batch of 
%foreground proposals 
$\{x_i, c_i\}_{i=1}^N$, where $c_i$ is the ground-truth label of the $i_{th}$ feature $x_i$ from the ROI feature extractor, the multi-level semantics-aware hierarchical contrastive loss (MSA-hicl) is computed by:
\begin{equation} 
   \begin{split}
   \mathcal{L}_{\mathrm{MSA-hicl}} = \frac{1}{N} \sum_{i=1}^{N} \left ( \frac{-1}{\sum_{l=0}^L \varphi(l)} \sum_{l=0}^{L} \varphi(l) \cdot  F(l) \right )
  \label{eq:10}
  \end{split}
\end{equation}
\begin{equation} 
   F(l) = \frac{x_i \cdot M_{c_i}^l}{\tau} - \log{}{\sum_{\substack{c_j \in C, \\ \hat{l} \in [0, L]}}}\left( s(N_{c_i}^l, N_{c_j}^{\hat{l}})\cdot exp(x_i \cdot M_{c_j}^{\hat{l}} / \tau -b)\right) -b
  \label{eq:11}
\end{equation}
where we use a Log-Sum-Exp (LSE) strategy to prevent the overflow problem. $\tau$ is the temperature hyperparameter, $\varphi(\cdot)$ is the aggregation strength of different levels, and $b$ is the maximum value in {$\{x_i \cdot M_{c_j}^{\hat{l}}/\tau\}$}. {$s(N_{c_i}^l, N_{c_j})$} is the similarity between two nodes, which is computed by 
\begin{equation} 
    s(N_{c_i}^l, {N_{c_j}^{\hat{l}}})= exp(-R(W_{N_{c_i}^l}, W_{N_{c_j}^{\hat{l}}})) 
  \label{eq:12}
\end{equation}
where {\scriptsize $W_{N_{c_i}^l}$} and {\scriptsize $W_{N_{c_j}^{\hat{l}}}$} are semantic vectors of two nodes. For a leaf node {$N_{c_i}^L$},  {\scriptsize $W_{N_{c_i}^L}$} is the semantic vectors of corresponding category, and for a non-leaf node {$N_{c_i}^l$}, {\scriptsize $W_{N_{c_i}^l}$} aggregates the semantic vectors from its leaf node descendants.

In Eq.~(\ref{eq:10}), we uses aggregation weights $\varphi(\cdot)$ to narrow the distance between $x_i$ and {$\{\theta_{c_i}^l|l \in [0, L]\}$}, and utilize similarity weights $s$ in Eq.~(\ref{eq:11}) to align semantic distances to {$\{\theta_{C^{\setminus c_i}}^l|l \in [0, L]\}$}. Therefore, we can cluster subclasses under the parent node, reducing intra-class variance while increasing inter-class separability.

\subsubsection{Momentum update strategy.} In order to make the parameter changes of the momentum cache smoother and more stable, we propose a momentum update strategy to continuously refine these representations.  Specifically, for a feature map $x_i$ with a ground-truth label $c_i$, the updates for all {$\{\theta_{c_i} ^l|l \in [0,L]\}$} computed as
\begin{equation} 
    \theta_{c_{i}}^{l}\, \leftarrow m\cdot \theta_{c_{i}}^{l} + (1-m)\cdot x_{i} \, , \ \ {l \in [0,L]}
  \label{eq:14}
\end{equation}
where $m \in [0, 1)$ is a momentum coefficient and set to 0.99 by default. We initialize these momentum caches using aligned semantic vectors in Eq.~(\ref{eq:5}), which provides a reliable starting point for momentum updates. In this way, it maintains semantically rich, hierarchically structured momentum caches and prevents potential drift during training.

\subsection{Training Loss}
The total loss of our model used for training is as follows:
\begin{equation} 
\begin{split}
   \mathcal{L}_{total} &= \mathcal{L}_{RPN_{reg/cls}} +  \mathcal{L}_{Reg} + \lambda_1\mathcal{L}_{avss} \\
   &\quad + \lambda_2(\mathcal{L}_{D} + \mathcal{L}_{G}) + \lambda_3\mathcal{L}_{MSA-hicl} 
\end{split}
\label{eq:16}
\end{equation}
where $\mathcal{L}_{RPN_{reg/cls}}$ is the classification and regression loss of RPN, $\mathcal{L}_{Reg}$ is the regression loss in RoI head, and $\lambda_1$, $\lambda_2$, and $\lambda_3$ control the weights of the last three losses.

% \begin{table}[t]
%   \centering
%   \caption{FG-ZSD performance for unseen classes of Recall@100 and mAP with different IoU thresholds.}
%   % \vspace{-0.5em}
%   \resizebox{0.44\textwidth}{!}{
%   \begin{tabular}{@{}ccccc@{}}
%     \toprule
%     \multirow{2}{*}{Method} & \multicolumn{3}{c}{Recall@100} & mAP\\
%     \cmidrule(lr){2-4}
%     & IoU=0.4 & IoU=0.5 & IoU=0.6 & IoU=0.5\\
%     \midrule
%     ZSIS \cite{zheng2021zero}& 0.003 & 0.003 & 0.001 & 0.001 \\
%     RRFS \cite{gen3}& 0.007 & 0.006 & 0.005 & 0.005 \\
%     SCR \cite{sarma2022resolving}& 0.005 & 0.004 & 0.003 & 0.004 \\
%     SeeDs \cite{gen5}& 0.007 & 0.006 & 0.005 & 0.005 \\
%     TCB~\cite{li2023zero}& 0.004 & 0.002 & 0.002 & 0.003 \\
%     SAUI~\cite{wang2024saui} & 0.008 & 0.007 & 0.006 & 0.006  \\
%     M-RRFS~\cite{huang2024m}&0.01 & 0.009 & 0.008 & 0.008  \\ 
%     \textbf{MSHC~(Ours)}& \textbf{24.4}& \textbf{24.3}& \textbf{24.3}& \textbf{8.046} \\
%     \bottomrule
%   \end{tabular}
%     }
%   \label{tab:3}
% \end{table}

\begin{table}[t]
  \centering
   
  \caption{FG-ZSD performance for unseen classes of Recall@100 and mAP with different IoU thresholds.}
  % \vspace{-0.5em}
  \resizebox{0.33\textwidth}{!}{
  \begin{tabular}{@{}ccccc@{}}
    \toprule
    \multirow{2}{*}{Method} & \multicolumn{3}{c}{Recall@100} & mAP\\
    \cmidrule(lr){2-4}
    & IoU=0.4 & IoU=0.5 & IoU=0.6 & IoU=0.5\\
    \midrule
    ZSIS \cite{zheng2021zero}& 4.23 & 4.23 & 4.21 & 1.45 \\
    RRFS \cite{gen3}& 4.57 & 4.57 & 4.55 & 1.66 \\
    SCR \cite{sarma2022resolving}& 4.63 & 4.63 & 4.61 & 1.66 \\
    SeeDs \cite{gen5}& 6.73 & 6.73 & 6.73 & 2.33 \\
    TCB~\cite{li2023zero}& 7.65 & 7.65 & 7.63 & 2.97 \\
    SAUI~\cite{wang2024saui} & 8.24 & 8.23 & 8.21 & 3.25  \\
    M-RRFS~\cite{huang2024m}& 9.01 & 9.01 & 8.99 & 4.11 \\ 
    \textbf{MSHC~(Ours)}& \textbf{24.4}& \textbf{24.3}& \textbf{24.3}& \textbf{11.4} \\
    \bottomrule
  \end{tabular}
    }
  \label{tab:3}
\end{table}

\section{Experiments}
\label{6_experiment}
This section presents extensive experiments to demonstrate the effectiveness of the MSHC model. 
% Due to the space limit, we provide additional experimental settings in the supplementary material.

\subsection{Implementation Details}
The object detection module uses Faster-RCNN~\cite{girshick2015fast} with ResNet-101~\cite{he2016deep} pretrained on ImageNet~\cite{russakovsky2015imagenet}. 
%as the backbone,
The generative network consists of several fully connected layers with MLPs and ReLU layers, and pretrained on CUB~\cite{wah2011caltech}.
The hyperparameter of $\xi$ in Eq.~(\ref{eq:4}) is set to 5, $\tau$ in Eq.~(\ref{eq:11}) is set to 0.5, and the aggregation function $\varphi(l) = l$. 
The hyperparameters $\lambda_1$, $\lambda_2$ and $\lambda_3$ in Eq.~(\ref{eq:16}) are set to 0.1, 0.5 and 0.01, respectively. Due to space limit, ablation studies on $\varphi(l)$ and other hyperparameters
%In Sec.~\ref{sec:Ablation Study}, we present the experimental results on how these parameters impact the performance. 
are available along with the dataset and code in the 
supplementary material.

\subsection{Performance in FG-ZSD Setting} 
We compare the proposed method with existing SOTA ZSD methods under the FG-ZSD setting, as presented in Tab.~\ref{tab:3}. Given the demand for capturing subtle class differences in FG-ZSD, 
almost all existing ZSD methods accurately predict unseen classes at the fine-grained level with extremely low probability due to their heavy reliance on global image features and generic class attributes. 
Our method outperforms all the compared existing methods on all metrics, demonstrating its effectiveness and superiority in FG-ZSD. Specifically, it significantly outperforms the SOTA method M-RRFS~\cite{huang2024m} in terms of mAP and Recall@100 for all IoU thresholds, improving Recall@100 by 15.29\% and mAP by 7.29\% at IoU=0.5.

\subsection{Performance in FG-GZSD Setting}
We conduct performance comparison under the FG-GZSD setting, which is a more challenging and realistic setting. The results are in Tab.~\ref{tab:4}. Our method achieves significant improvements on both seen and unseen classes, with the HM of mAP improving from 5.8\% to 17.5\%, and Recall@100 improving by 10.6\% and 14.98\% for seen and unseen classes, respectively. To further demonstrate the effectiveness, we also report the performance of seen classes on all metrics in Tab.~\ref{tab:5}, where at IoU=0.5, our method improves Recall@100 from 79.7\% to 91.1\%, as well as mAP from 71.5\% to 78.5\% respectively.

\begin{table}[t]
\scriptsize
  \centering
  \caption{Comparison of Recall@100 and mAP on seen~(S) and unseen~(U) classes at IoU=0.5 under FG-GZSD setting. HM: the harmonic mean to assess a model's combined performance on S and U. These notations are used in other tables.}
  \resizebox{0.36\textwidth}{!}{
  \begin{tabular}{@{}ccccccc@{}}
    \toprule
    \multirow{2}{*}{Method} & \multicolumn{2}{c}{Recall@100} &\multicolumn{3}{c}{mAP}\\
    \cmidrule(lr){2-3} \cmidrule(lr){4-6}
    & S & U & S & U & HM \\
    \midrule
    ZSIS~\cite{zheng2021zero}& 77.9 & 3.13 & 69.7 & 1.3 & 2.552\\
    RRFS~\cite{gen3}& 79.5 & 3.36 & 70.3 & 1.53 & 2.995 \\
    SCR~\cite{sarma2022resolving}& 77.5 & 3.99 & 71.4 & 1.54 & 3.015 \\
    SeeDs~\cite{gen5}& 78.7 & 5.78 & 69.3 & 2.04 & 3.963\\
    TCB~\cite{li2023zero}& 78.2 & 6.15 & 69.0 & 2.55 & 4.918\\
    SAUI~\cite{wang2024saui}& 79.7 & 7.68 & 71.3 & 2.88 & 5.536\\
    M-RRFS~\cite{huang2024m}& 80.5 & 8.22 & 71.7 & 3.03 & 5.814\\
    \textbf{MSHC~(Ours)}& \textbf{91.1} & \textbf{23.2} & \textbf{78.5} & \textbf{\textbf{9.85}
} & \textbf{17.504}\\
    \bottomrule
  \end{tabular}
    }
  \label{tab:4}
\end{table}

\begin{table}[t]
  \centering
  \caption{FG-GZSD performance of seen classes in terms of Recall@100 and mAP with different IoU thresholds.}
  \resizebox{0.33\textwidth}{!}{
  \begin{tabular}{@{}ccccc@{}}
    \toprule
    \multirow{2}{*}{Method} & \multicolumn{3}{c}{Recall@100} & mAP\\
    \cmidrule(lr){2-4}
    & IoU=0.4 & IoU=0.5 & IoU=0.6 & IoU=0.5\\
    \midrule
    ZSIS~\cite{zheng2021zero}& 78.0 & 77.9 & 77.8 & 69.7 \\
    RRFS~\cite{gen3}& 79.6 & 79.5 & 79.1 & 70.3 \\
    SCR~\cite{sarma2022resolving}& 77.6 & 77.5 & 77.2 & 71.4 \\
    SeeDs~\cite{gen5}& 78.8 & 78.7 & 78.2 & 69.3 \\
    TCB~\cite{li2023zero} & 78.1 & 78.0 & 77.7 & 70.7 \\
    SAUI~\cite{wang2024saui} & 79.3 & 79.2 & 78.9 & 70.5 \\
    M-RRFS~\cite{huang2024m}& 79.8 & 79.7 & 79.4 & 71.5 \\ 
    \textbf{MSHC~(Ours)}& \textbf{91.2} & \textbf{91.1} & \textbf{90.6} & \textbf{78.5}\\
    \bottomrule
  \end{tabular}
  }
  \label{tab:5}
\end{table}

\begin{table}[t]
    \centering
    \caption{Effects of different components of mAP at IoU=0.5.}
    % \vspace{-0.5em}
    \resizebox{0.36\textwidth}{!}{
        \begin{tabular}{@{}cccccccc@{}}
        \toprule
        \multirow{2}{*}{Method} & \multirow{2}{*}{MSA-GN} & \multirow{2}{*}{MSA-hicl} & \multirow{2}{*}{FG-ZSD} & \multicolumn{3}{c}{FG-GZSD} \\ 
        \cmidrule(lr){5-7}
        &     &     &     & S    & U    & HM   \\ 
        \midrule
        A &     &      & 8.4 & 73.4 & 5.8 & 10.751 \\
        B &  & \checkmark & 9.6 & 77.5 & 6.69 & 12.317 \\
        C & \checkmark &     & 10.8 & 75.2 & 8.55 & 15.354 \\ 
        D & \checkmark & \checkmark & \textbf{11.4} & \textbf{78.5} &  \textbf{9.85} & \textbf{17.504} \\
        \bottomrule
\end{tabular}
}
\label{tab:6}
\end{table}

\subsection{Ablation Studies} 
\subsubsection{Effects of major modules}
We check the effects of two major modules MSA Generative Network (MSA-GN) and MSA-hicl loss by considering four configurations: 1) the baseline (the 1st row) using a hierarchical Faster-RCNN, Eq.~(\ref{eq:7}) to perform affine transformations,   and CE loss to train the classifier.
2) Without MSA-GN, using only Eq.~(\ref{eq:7}) to perform affine transformations. 3) Without MSA-hicl loss, using CE loss to train the classifier. 4) Our method. The results are reported in Tab.~\ref{tab:6}. We can see that 
the MSA-hicl loss notably boosts mAP for seen and unseen classes, from 73.4\% to 77.5\% and 5.8\% to 6.69\% respectively. MSA-GN improves mAP from 73.4\% to 75.2\% for seen classes and from 5.8\% to 8.55\% for unseen classes. 
Combining MSA-GN and MSA-hicl, our model achieves the highest mAP gain, up to 4.05\% for unseen classes in FG-ZSD, and attains the best HM in FG-GZSD, proving the effectiveness of our method.

\subsubsection{Ablation study on aggregation function} 
We analyze the aggregation function $\varphi(l)$, which narrows the distance between feature map $x_i$ and hierarchical momentum caches ${\theta_{c_i}^l}_{l=0}^L$ for all nodes on the path of $c_i$ in the hierarchical tree $H_t$. Results in Tab.~\ref{table:1} reveal optimal performance with $\varphi(l)=l$. Equal weights ($\varphi(l)=1$) degrade mAP by 1.1\% (seen) and 1.63\% (unseen) due to inadequate hierarchical differentiation. Conversely, quadratic weighting ($\varphi(l)=l^2$) yields more compact subcategories and broader parent categories. This enhances the nuanced distinctions between categories, leading to improved model performance.
\begin{table}[t]\footnotesize
    \centering
    \caption{Results of ablation study on function $\varphi(\cdot)$.}
    \resizebox{0.33\textwidth}{!}{
    \begin{tabular}{@{}cccc@{}}
    \toprule
    \multirow{2}{*}{Aggregation Function \( \varphi(\cdot) \)} & \multicolumn{3}{c}{mAP} \\ 
    \cmidrule(l){2-4}
    & S & U & HM \\
    \midrule
    $\varphi(l) = 1$ & 77.4 & 8.22 & 14.862 \\
    $\varphi(l) = l$ & \textbf{78.5} & \textbf{9.85} & \textbf{17.504} \\
    $\varphi(l) = l^2$ & 78.0 & 9.62 & 17.127 \\
    \bottomrule
    \end{tabular}
    }
\label{table:1}
\end{table}

\begin{table}[t]\footnotesize
    \centering
    \caption{Results of ablation study on hyperparameters.}
    \resizebox{0.3\textwidth}{!}{
    \begin{tabular}{@{}cccc@{}}
    \toprule
    Hyperparameter setting & S & U & HM \\
    \midrule
    $\xi=0.1,\ \tau=0.5$ & 78.1 & 8.74 & 15.721 \\
    $\xi=1,\ \tau=0.5$ & 78.2 & 9.43 & 16.83 \\
    $\xi=5,\ \tau=0.5$  & \textbf{78.5} & \textbf{9.85} & \textbf{17.504} \\
    $\xi=10,\ \tau=0.5$  & 78.4 & 9.82 & 17.454 \\
    \cmidrule(l){1-4}
    $\xi=5,\ \tau=0.05$  & 77.9 & 9.55 & 17.014 \\
    $\xi=5,\ \tau=0.2$  & 78.3 & 9.63 & 17.151 \\
    $\xi=5,\ \tau=1$  & 77.5 & 8.82 & 15.838 \\
    \bottomrule
    \end{tabular}
    }
\label{table:2}
\end{table}

\subsubsection{Hyperparameters Sensitivity}
We investigate temperature hyperparameters $\xi$ (Eq.~(\ref{eq:4})) and $\tau$ (Eq.~(\ref{eq:11})). Tab.~\ref{table:2} indicates optimal results with $\tau=0.5$ and increasing mAP for both seen and unseen classes as $\xi$ grows, with the best performance at $\xi=5$. This indicates that appropriately increasing $\xi$ helps generate high-quality and fine-grained images based on the given text description. The reason is that the proposed attention-based visual-semantic similarity loss can promote the generator to focus on the words related to the image.

\subsubsection{Visualization effect of MSA-hicl}
\begin{figure}[h]
  \centering
   \includegraphics[width=0.9\linewidth]{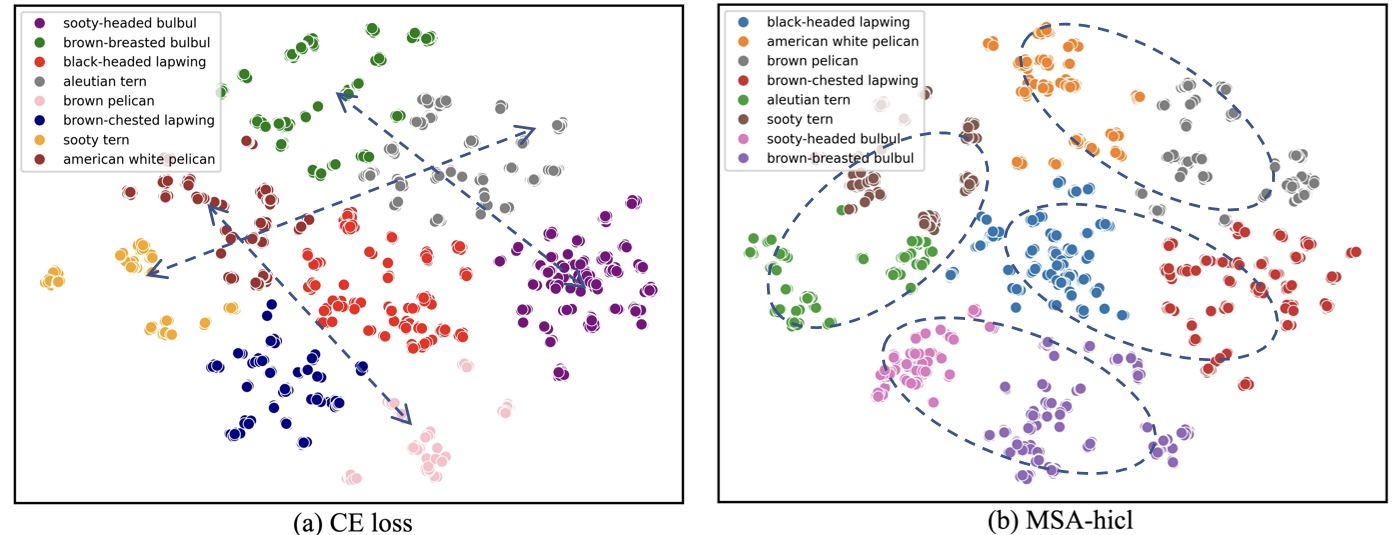}
   \caption{t-SNE visualization of box features. The feature space learned (a) by the CE loss is chaotic, while (b) by MSA-hicl is distinctive and clearly organized by genus.}
   \label{fig:tsne}
\end{figure}

% Fig.~\ref{fig:tsne} illustrates the visualization comparison of the feature space learnt by (a) cross-entropy (CE) loss and (b) our proposed multi-level semantics-aware hierarchical contrast loss (MSA-hicl). To ensure clarity and visual comprehensibility, we select eight species from four genera.
% CE loss feature space blends species, showing its weakness in distinguishing similar species. Conversely, MSA-hicl feature space is more distinct and organized by genus, reflecting both coarse-grained and fine-grained semantic distinctions, thus improving the model's discrimination and generalization.
Fig.~\ref{fig:tsne} visualizes feature embeddings trained with (a) CE loss and (b) MSA-hicl loss, using eight species across four genera for clarity. CE loss embeddings show significant overlap among similar species. In contrast, MSA-hicl embeddings distinctly cluster by genus, demonstrating improved semantic differentiation at both coarse and fine-grained levels, thereby enhancing model's discriminative power and generalization.

\section{Conclusion}
In this paper, we introduce Fine-Grained Zero-Shot Object Detection (FG-ZSD), a nontrivial extension to the existing ZSD task. To solve FG-ZSD, we construct FGZSD-Birds, a large-scale, high-quality dataset containing 148,820 images across 1,432 categories, supporting FG-ZSD and various related tasks. 
Additionally, we develop the first FG-ZSD method MSHC, which uses multi-level aligned semantic vectors to generate fine-grained images of unseen classes, and multi-level semantics-aware hierarchical contrastive learning to constrain the feature space. Extensive experiments show that our method outperforms existing ZSD models.

\section{Acknowledgment}
Jihong Guan was supported by National Natural Science Foundation of China (NSFC) under grants No. 62172300 and No. 62372326. Jiaogen Zhou was supported by NSFC under grant No. 32470441. The computations in this research were performed using the CFFF platform of Fudan University.

% In this paper, we propose a new problem called fine-grained zero-shot object detection (FG-ZSD in short), which is a nontrivial extension to the existing ZSD task. To solve FG-ZSD, on the one hand, we construct the first large-scale, high-quality FG-ZSD dataset FGZSG-birds, which contains 148,820 images and 1,432 categories and supports not only FG-ZSD but also dozen of other tasks. On the other hand, we develop the first FG-ZSD method MSHC, which uses multi-level aligned semantic vectors to generate fine-grained images of unseen class, and multi-level semantics-aware hierarchical contrastive learning to constrain the feature space. Extensive experiments on FGZSD-Birds show that our method outperforms existing ZSD models.
\bibliographystyle{ACM-Reference-Format}
\bibliography{main}

\end{document}